\IfFileExists{./prepreamble-amspreprint.sty}{\RequirePackage[packages,theorems,changes]{prepreamble-amspreprint}}{}

\makeatletter
\IfFileExists{./scoop-latex/scoop-packages.sty}{\providecommand*{\input@path}{}\edef\input@path{{./scoop-latex/}\input@path}}{}
\makeatother
\PassOptionsToPackage{sorting = none, style = nature, backend = biber}{biblatex}
\PassOptionsToPackage{commentmarkup = footnote}{changes}
\PassOptionsToPackage{ngerman, english}{babel}    

\documentclass[review,anonymous,format = sigconf]{amsart}

\RequirePackage{biblatex}
\RequirePackage{ltxcmds}
\IfFileExists{./preamble-amspreprint.sty}{\RequirePackage[packages,theorems,changes]{preamble-amspreprint}}{}

\usepackage{scoop-local}

\IfFileExists{./postpreamble-amspreprint.sty}{\RequirePackage[packages,theorems,changes]{postpreamble-amspreprint}}{}

\makeatletter
\@ifpackageloaded{changes}{
\definechangesauthor[name = {Viktor Martinek}, color = {red!80!black}]{VM}
\definechangesauthor[name = {Roland Herzog}, color = {blue!80!black}]{RH}
}{}
\makeatother        

\makeatletter
\@ifpackageloaded{hyperref}{%
	\hypersetup{
		pdftitle = {Symbolic Regression for Shared Expressions},
		pdfauthor = {Viktor Martinek, Roland Herzog},
		pdfkeywords = {symbolic regression, symbolic machine learning, multi-view symbolic regression, class symbolic regression, factor variables},
		pdfcreator = {Created using the Scoop Template Engine version 1.6.0.}
	}
}{
	\pdfinfo{
		/Title (Symbolic Regression for Shared Expressions)
		/Author (Viktor Martinek, Roland Herzog)
		/Subject ()
		/Keywords (symbolic regression, symbolic machine learning, multi-view symbolic regression, class symbolic regression, factor variables)
		/Creator (Created using the Scoop Template Engine version 1.6.0.)
	}
}
\makeatother

\title[Symbolic Regression for Shared Expressions]{Symbolic Regression for Shared Expressions \\ Introducing Partial Parameter Sharing}

\author[V. Martinek]{Viktor Martinek\orcidlink{0000-0001-6215-4783}}
\address[V. Martinek]{Interdisciplinary Center for Scientific Computing, Heidelberg University, 69120 Heidelberg, Germany}
\email{\detokenize{viktor.martinek@iwr.uni-heidelberg.de}}

\author[R. Herzog]{Roland Herzog\orcidlink{0000-0003-2164-6575}}
\address[R. Herzog]{Interdisciplinary Center for Scientific Computing, Heidelberg University, 69120 Heidelberg, Germany}
\email{\detokenize{roland.herzog@iwr.uni-heidelberg.de}}

\thanks{This work was funded by the Deutsche Forschungsgemeinschaft (DFG, German Research Foundation) within the Priority Programme SPP~2331 (Machine Learning in Chemical Engineering) – Project number~466528284.
TiSR is the result of a joint effort by Viktor Martinek and Roland Herzog at Heidelberg University, as well as Ophelia Frotscher and Markus Richter at Leibniz University Hannover.}

\date{\today}

\dedicatory{}

\begin{document}

% Insert the abstract.
\begin{abstract}
%! TEX root = ./manuscript-llncs.tex

\Acl{SR} aims to find symbolic expressions that describe datasets.
Due to its inherent interpretability, \ac{SR} is a powerful paradigm for scientific discovery.
Recent advances have expanded \ac{SR} to describe related phenomena using a single expression with varying sets of parameters, thereby introducing a single categorical variable.
To illustrate, this enables the search for a single expression describing temperature-dependent viscosity across multiple fluids, while simultaneously identifying a distinct set of fluid-specific parameters.
We expand upon prior efforts by considering multiple categorical variables and introducing intermediate levels of parameter sharing.
Rather than parameters being either entirely universal or entirely unique, some parameters can also be shared across specific categories while remaining distinct for others.
This allows for separating universal effects (shared parameters), category-specific trends (partially-shared parameters), and category interactions (non-shared parameters).
We test the limits of this setup in terms of reducing data requirements and transfer learning using a synthetic, fitting-only example.
Furthermore, we apply the method to an astrophysics dataset also used in a previous single-category study.
In comparison, we achieve similar fit quality with significantly fewer parameters while extracting additional information about the problem.

\end{abstract}

% Insert the keywords.

% Insert the Mathematics Subject Classification.
\makeatletter
\ltx@ifpackageloaded{hyperref}{%
\subjclass[2010]{}
}{%
\subjclass[2010]{}
}
\makeatother

% Typeset the opening page.
\maketitle

% Insert the document body.
%! TEX root = ./manuscript-llncs.tex

\acresetall

\section{Introduction}
\label{section:introduction}

\Ac{SR} is a machine learning paradigm that distills symbolic expressions from data.
Other machine learning methods assume a fixed functional form and seek to identify its parameters.
In contrast, \ac{SR} searches for both the functional form and its associated parameters, aiming for more concise representations.
Symbolic expressions are human interpretable and have long been used to describe and study phenomena in various disciplines.
As such, \ac{SR} is a uniquely powerful tool for scientific discovery, system identification, and explainable artificial intelligence.

Koza~\cite{Koza:1992:1} introduced \ac{SR} itself and proposed an evolutionary-inspired method, \ie \ac{GP}.
Putting \ac{GP} in simple terms, a population of randomly generated expressions is randomly combined and changed.
Inferior expressions are discarded in favor of better ones, as determined by the optimization objectives that balance fit quality and expression complexity.
Beyond \cite{Koza:1992:1}, \ac{GP} methods have been continually improved in the last $30$~plus years and are still among the best approaches to \ac{SR} \cite{LaCavaOrzechowskiBurlacuOlivettiDeFrancaVirgolinJinKommendaMoore:2021:2,ImaiAldeiaZhangBomaritoCranmerFonsecaBurlacuLaCavaOlivettiDeFranca:2025:2}.
There are numerous other approaches for \ac{SR} including, but not limited to, exhaustive search \cite{KammererKronbergerBurlacuWinklerKommendaAffenzeller:2020:1,BartlettDesmondFerreira:2024:1}, reinforcement-learning-based methods \cite{Michishita:2024:1,PetersenLandajuelaLarmaMundhenkPrataSantiagoKimKim:2021:1,LandajuelaLeeYangGlattSantiagoAravenaMundhenkMulcahyPetersen:2022:1}, and amortized approaches \cite{BiggioBendinelliNeitzLucchiParascandolo:2021:1,ValipourYouPanjuGhodsi:2021:1}.
Although some are well-poised to do so, these have not yet eclipsed \ac{GP}-based methods in all respects.

\Ac{SR} has successfully been applied to scientific discovery in many domains, \eg, astrophysics \cite{MatchevMatchevaRoman:2022:1,CranmerSanchezGonzalezBattagliaXuCranmerSpergelHo:2020:1,LemosJeffreyCranmerHoBattaglia:2023:1}, material science \cite{WangWagnerRondinelli:2019:1,WangWangLiZhouSun:2024:1,BurlacuKommendaKronbergerWinklerAffenzeller:2023:1}, and thermodynamics \cite{MuznyHuberKazakov:2013:1,FrotscherMartinekFingerhutYangVrabecHerzogRichter:2023:1,YangFrotscherRichter:2025:1,SotiriadouAntoniadisAssaelMartinekHuber:2025:1,YangRichter:2025:1}.
For some problems, it is beneficial to describe similar systems using the same functional form and allow it to differ only in the parameter values.
For example, \cite{MartinekBellHerzogRichterYang:2025:1} developed a single, \emph{shared} expression linking the residual entropy to the viscosity of $124$~fluids.
Some of the parameters are fluid-agnostic and are \emph{shared} among all fluids, while other parameters are fluid-specific (\emph{non-shared}) and change to accommodate the different behavior.
Intuitively, finding a functional form capable of describing multiple similar phenomena increases the chances of it being meaningful and of generalizing beyond the provided category values.
The chances of the functional form that is valid for $124$~fluids being valid for a $125^{\textup{th}}$~fluid are high.
Notably, this viscosity correlation was developed without the use of \ac{SR}.
This approach essentially introduces one categorical variable, with the various fluids serving as its different category values.\footnote{In this work, we use the term \enquote{categorical variables} in favor of \enquote{nominal variables}, as they allow simpler phrasing and thus clearer understanding. We do not imply any ordering of the category values.}

Thus, categorical variables allow different continuous datasets to be combined, either when describing the same phenomenon measured under different conditions or when describing different but similar phenomena.
In some cases, categorical variables may hide unwanted complexity or compress not-yet-understood interrelations of missing independent variables capable of explaining the distinct categorical values, as is also discussed in \cite{KronbergerKommendaPrombergerNickel:2018:1}.
For example, the ideal gas law is $p = \rho \frac{R}{M}T$, where $R$ is the ideal gas constant, $M$ is the substance-specific molar mass, and $p$, $\rho$, $T$ are the variables pressure, density, and temperature.
If the substance-specific molar masses were unknown, we could develop a shared expression defining each substance as a category value.
For the viscosity correlation mentioned, it remains unclear which latent continuous variables account for the variance between different fluids.

Sharing a functional form across category values regularizes the functional form \cite{RusseilOlivettiDeFrancaMalanchevBurlacuIshidaLerouxMichelinMoinardGangler:2024:1}.
This may prevent \enquote{symbolic overfitting}, improve generalization performance, and increase robustness to noise.
Some of the prior approaches use a fully distinct set of parameters for each category value, while others incorporate category-agnostic or \enquote{shared} parameters.
Sharing some of the parameters between the category values reduces the number of parameters and potentiates all mentioned benefits, as discussed in \cite{RusseilOlivettiDeFrancaMalanchevMoinardCherrey:2025:1}.
Furthermore, this approach also leads to transfer learning to some degree, where knowledge is reused across phenomena \cite{WeissKhoshgoftaarWang:2016:1,LohSchneegassTian:2019:1}.

In recent years, several works have explored incorporating a single categorical variable into \ac{SR} in such a (or a similar) manner with some differences in their approaches \cite{LohSchneegassTian:2019:1,TenachiIbataFrancoisDiakogiannis:2024:1,RusseilOlivettiDeFrancaMalanchevBurlacuIshidaLerouxMichelinMoinardGangler:2024:1,KronbergerKommendaPrombergerNickel:2018:1,RusseilOlivettiDeFrancaMalanchevMoinardCherrey:2025:1}; see also the implementations \pysr~\cite{Cranmer:2020:1} and \eggp~\cite{OlivettiDeFrancaKronberger:2025:2}.
However, no prior work considered two or more categorical variables, or proposed a method capable of taking the effects of multiple categorical variables and their interactions into account separately.

Problems with two or more categorical variables can be approached using category-value combinations as a single, condensed categorical variable.
However, this ignores a crucial part of the problem and fails to fully exploit the potential for parameter sharing and reduction.
To the best of our knowledge, all prior works but one have explored approaches capable of incorporating only one categorical variable.
Only Kronberger~\etal~\cite{KronbergerKommendaPrombergerNickel:2018:1} could, in principle, handle multiple categorical variables separately.
However, their approach cannot account for the interactions.
In the present work, we demonstrate a novel approach that uses three types of parameter sharing, \ie \enquote{fully-shared} (global), \enquote{partially-shared} (category-value-specific), and \enquote{non-shared} (category-value-combination-specific).

The proposed approach is introduced in the following section.
In \cref{section:related-work}, we discuss existing methods and compare them to our approach.
Finally, the approach and its benefits are demonstrated in \cref{section:application}, before a conclusion is drawn in \cref{section:conclusion}.

\section{Method}
\label{section:method}

In this section, for simplicity's sake, the proposed approach is shown for a two-category example.
However, the approach is applicable to arbitrarily many categorical variables.

\subsection{Concept}

To explain the concept, we introduce an example problem.
The problem has two categorical variables $U$ (uppercase) and $L$ (lowercase).
Category~$U$ has four category values (\texttt{A}, \texttt{B}, \texttt{C}, \texttt{D}), while Category~$L$ has three values (\texttt{a}, \texttt{b}, \texttt{c}).
Thus, in the datasets, there are four times three possible category-value combinations.

The core of our approach is the use of three distinct levels of parameter sharing: fully-shared, partially-shared, and non-shared, which depend on neither, either, or both category values, respectively.
The highest level of sharing is represented by fully-shared parameters, which are the same across all category-value combinations.
Non-shared parameters are unique to each category-value combination.
The intermediate sharing level is represented by partially-shared parameters.
These parameters are shared across one of the two categories, but depend on the values of the other category.
This approach allows us to distinguish category-agnostic effects, effects of either category, and also category-value-combination interactions.
\Cref{figure:example1} illustrates the sharing concepts for the example at hand.

\begin{figure}[htb]
	\centering
	% ==================================================================================================
% This code was first generated by chat-gpt at 2025-12-04-08h, and then adapted
% ==================================================================================================

\makeatletter
\newcommand{\MyResizeWidth}{\@ifclassloaded{acmart}{\linewidth}{0.6\textwidth}}
\makeatother
\resizebox{\MyResizeWidth}{!}{%
    \begin{tikzpicture}[]

        % ------------------------------------------------------------
        % GLOBAL SPACING VARIABLES
        % ------------------------------------------------------------
        % Inner grid spacing
        \def\ColW{1.0}    % horizontal distance between columns
        \def\RowH{1.0}    % vertical distance between rows

        % ------------------------------------------------------------
        % COLUMN AND ROW POSITIONS (derived automatically)
        % ------------------------------------------------------------
        \def\xA{1*\ColW}
        \def\xB{2*\ColW}
        \def\xC{3*\ColW}
        \def\xD{4*\ColW}

        \def\yA{-1*\RowH}
        \def\yB{-2*\RowH}
        \def\yC{-3*\RowH}

        % Quadrant spacing
        \def\Hsep{5*\ColW}      % horizontal separation between left ↔ right
        \def\Vsep{4*\RowH}      % vertical separation between top ↔ bottom

        % ------------------------------------------------------------
        % MACRO TO DRAW A SINGLE GRID (letters only)
        % ------------------------------------------------------------
        \newcommand{\PlaceGrid}{
            % Column labels
            \node at (\xA,0) {\texttt{A}};
            \node at (\xB,0) {\texttt{B}};
            \node at (\xC,0) {\texttt{C}};
            \node at (\xD,0) {\texttt{D}};

            % Row labels
            \node at (0,\yA) {\texttt{a}};
            \node at (0,\yB) {\texttt{b}};
            \node at (0,\yC) {\texttt{c}};

            % Entries
            \foreach \col/\x in {\texttt{A}/\xA, \texttt{B}/\xB, \texttt{C}/\xC, \texttt{D}/\xD} {
                \foreach \row/\y in {\texttt{a}/\yA, \texttt{b}/\yB, \texttt{c}/\yC} {
                    \node at (\x,\y) {\col\row};
                }
            }
        }

        % ------------------------------------------------------------
        % TOP LEFT QUADRANT (red rounded boxes around each cell)
        % ------------------------------------------------------------
        \begin{scope}[shift={(0,0)}]
            % Column/row labels + text
            \PlaceGrid

            % Add red boxes
            \foreach \col/\x in {\texttt{A}/\xA, \texttt{B}/\xB, \texttt{C}/\xC, \texttt{D}/\xD} {
                \foreach \row/\y in {\texttt{a}/\yA, \texttt{b}/\yB, \texttt{c}/\yC} {
                    \node[draw=red, rounded corners=6pt, inner sep=3pt]
                    at (\x,\y) {\col\row};
                }
            }
        \end{scope}

        % ------------------------------------------------------------
        % TOP RIGHT QUADRANT (one large green box)
        % ------------------------------------------------------------
        \begin{scope}[shift={(\Hsep,0)}]
            \PlaceGrid

            \node[
                draw=green!70!black,
                rounded corners=6pt,
                fit={(0.8*\ColW, -0.8*\RowH) (4.2*\ColW, -3.2*\RowH)}
            ] {};
        \end{scope}

        % ------------------------------------------------------------
        % BOTTOM LEFT QUADRANT (blue column-group boxes)
        % ------------------------------------------------------------
        \begin{scope}[shift={(0,-\Vsep)}]
            \PlaceGrid

            % Blue vertical boxes around columns
            \foreach \x in {\xA, \xB, \xC, \xD} {
                \node[
                    draw=blue,
                    rounded corners=6pt,
                    fit={(\x+0.2*\ColW,-0.8*\RowH) (\x-0.2*\ColW,-3.2*\RowH)}
                ] {};
            }
        \end{scope}

        % ------------------------------------------------------------
        % BOTTOM RIGHT QUADRANT (orange row-group boxes)
        % ------------------------------------------------------------
        \begin{scope}[shift={(\Hsep,-\Vsep)}]
            \PlaceGrid

            % Orange horizontal boxes around rows
            \foreach \y in {\yA, \yB, \yC} {
                \node[
                    draw=orange,
                    rounded corners=6pt,
                    fit={(0.8*\ColW, \y-0.2*\RowH) (4.2*\ColW,\y+0.2*\RowH)}
                ] {};
            }
        \end{scope}

        % ------------------------------------------------------------
        % DOTTED SEPARATORS
        % ------------------------------------------------------------
        % This part was very painful -> at least 50 re-compilations until it
        % was visually pleasing...

        % horizontal dotted line (midway between top and bottom grids)
        \draw[dotted, thick]
          ({ -0.3*\ColW }, { -\Vsep + 0.3*\RowH }) -- ({ 2 * \Hsep - 0.5 * \ColW }, { -\Vsep + 0.3*\RowH});

        % vertical dotted line (midway between left and right grids)
        \draw[dotted, thick]
          ({ \Hsep - 0.4 * \ColW }, { -2*\Vsep + 0.5 * \RowH}) -- ({ \Hsep - 0.4 * \ColW }, {0.2 * \RowH});

    \end{tikzpicture}
}
	\caption{%
		Illustrative example of a two-category problem with all possible category-value combinations.
		Non-shared parameters are illustrated in the top-left, fully-shared ones in the top-right, and partially-shared ones in the bottom two illustrations.
	}
	\label{figure:example1}
\end{figure}

As an intuitive example, if $U$ were patient groups and $L$ were medications, a parameter partially shared across the patient types may account for the effects of a particular medicine on all patient types.
A parameter partially shared across medications may account for a particular patient-type's reaction to medication in general.
Finally, non-shared parameters would correct for binary interactions of particular medications and particular patient types, \eg, allergies.
The approach is reminiscent of mixed-effect models or group-contribution methods.

\subsection{Evaluation and Parameter Identification}
\label{subsection:evaluation-and-parameter-identification}

Our implementation shares some similarity with Kronberger~\etal~\cite{KronbergerKommendaPrombergerNickel:2018:1}, with some extra considerations regarding the new types of parameter sharing.
Shared parameters $p_{\textup{shared}}$ consist of a single real number minimized across the entire dataset.
Partially-shared parameters depending on category~$U$, denoted as $p_{\textup{partial},U}$, contain one real value for each unique level $u \in U$; the same logic applies to other categories.
Non-shared parameters $p_{\textup{nonshared}}$ are optimized for each category-value combination $(u,l)$ individually, containing as many real numbers as there are unique category-value combinations $(u,l)$.

For each category pair $(u,l)$ and each data point $i$ belonging to that pair, the observed value $y_{u,l,i}$ is predicted with
\begin{equation}
	\label{eq:predict}
	\hat{y}_{u,l,i}
	=
	f\paren[big](){%
		x_{u,l,i};\,
		p_{\textup{shared}},\,
		p_{\textup{partial},U,u},\,
		p_{\textup{partial},L,l},\,
		p_{\textup{nonshared},(u,l)}
	}
	.
\end{equation}
The least-squares fitting objective is formulated as a nested loop over the categorical variables and the data points $\cI_{u,l}$ within each category-value combination:
\begin{equation}
	\label{eq:minimize}
	\text{Minimize}
	\quad
	\sum_{u \in U}
	\;
	\sum_{l \in L}
	\;
	\sum_{i \in \cI_{u,l}}
	\paren[big](){%
		y_{u,l,i}
		-
		\hat{y}_{u,l,i}
	}^{2}
	\quad
	\text{\wrt\ }
	P
	,
\end{equation}
where $P = \paren[big](){p_{\textup{shared}},\, p_{\textup{partial},U},\, p_{\textup{partial},L},\, p_{\textup{nonshared}}}$.

To illustrate, we introduce the shared expression
\makeatletter
\begin{multline}
	\label{eq:example1}
	\hat{y}_{u,l,i}
	=
	p_{\textup{shared}}          \cdot v_1
	+
	p_{\textup{partial},U,u}     \cdot v_1^2
	\@ifclassloaded{acmart}{\\}{}
	+
	p_{\textup{partial},L,l}     \cdot v_1^3
	+
	p_{\textup{nonshared},(u,l)} \cdot v_1^4,
\end{multline}
\makeatother
where
$v_1$ is a continuous, independent variable,
$p_{\textup{shared}} = 100$ is a fully-shared parameter,
$p_{\textup{partial},U} = \set{\texttt{A} \mapsto 10,\, \texttt{B} \mapsto 20,\, \texttt{C} \mapsto 30,\, \texttt{D} \mapsto 40}$ is a partially-shared parameter depending on the categorical variable $U$,
$p_{\textup{partial},L} = \set{\texttt{a} \mapsto 1,\, \texttt{b} \mapsto 2,\, \texttt{c} \mapsto 3}$ is a partially-shared parameter depending on $L$, and
$p_{\textup{nonshared}} = \set{\texttt{Aa} \mapsto 0.01,\, \texttt{Ab} \mapsto 0.02,\, \texttt{Ac} \mapsto 0.03,\, \texttt{Ba} \mapsto 0.04,\, \texttt{Bb} \mapsto 0.05,\, \texttt{Bc} \mapsto 0.06,\, \texttt{Ca} \mapsto 0.07,\, \texttt{Cb} \mapsto 0.08,\, \texttt{Cc} \mapsto 0.09,\, \texttt{Da} \mapsto 0.1,\, \texttt{Db} \mapsto 0.11,\, \texttt{Dc} \mapsto 0.12}$ is a non-shared parameter different for each category-value combination~$(u,l)$.

\Cref{table:example1} illustrates which values each parameter assumes depending on the category-value combination using \cref{eq:example1} for a simple dataset.
The sparsity pattern of the Jacobian (consisting of the partial derivatives of the predictions $\hat{y}_{u,l,i}$ \wrt the parameters~$P$) required for identifying the parameters of this example is shown in \cref{figure:sparsity}.
This sparsity pattern can be exploited during differentiation to avoid a time and memory complexity of $O(nk)$, where $n$ is the number of data points and $k$ denotes the number of \enquote{individual} parameters\footnote{We refer to the real-valued parameters contained in each of the \enquote{high-level} parameters as individual parameters.} ($1 + 4 + 3 + 12 = 20$ in this case), to achieve a time and memory complexity of $O(nm)$, where $m$ is the number of parameters ($1 + 1 + 1 + 1 = 4$ in this case).

\begin{table*}[htb]
	\centering
	\caption{Values assumed by the different parameter types across category-value combinations for the example in \cref{eq:example1} using a simple dataset.}
	\begin{tabular}{lllllllll}
		\toprule
		$u$          & $l$          & $i$ & $v_1$ & $p_{\textup{shared}}$ & $p_{\textup{partial},U,u}$ & $p_{\textup{partial},L,l}$ & $p_{\textup{nonshared},(u,l)}$ & $y$    \\
		\midrule
		\texttt{A}   & \texttt{a}   &  1  & 1     & 100                   & 10                         & 1                          & 0.01                           & 111.01 \\
		\texttt{A}   & \texttt{b}   &  1  & 1     & 100                   & 10                         & 2                          & 0.02                           & 112.02 \\
		\texttt{A}   & \texttt{c}   &  1  & 1     & 100                   & 10                         & 3                          & 0.03                           & 113.03 \\
		\texttt{B}   & \texttt{a}   &  1  & 1     & 100                   & 20                         & 1                          & 0.04                           & 121.04 \\
		\texttt{B}   & \texttt{b}   &  1  & 1     & 100                   & 20                         & 2                          & 0.05                           & 122.05 \\
		\texttt{B}   & \texttt{c}   &  1  & 1     & 100                   & 20                         & 3                          & 0.06                           & 123.06 \\
		\texttt{C}   & \texttt{a}   &  1  & 1     & 100                   & 30                         & 1                          & 0.07                           & 131.07 \\
		\texttt{C}   & \texttt{b}   &  1  & 1     & 100                   & 30                         & 2                          & 0.08                           & 132.08 \\
		\texttt{C}   & \texttt{c}   &  1  & 1     & 100                   & 30                         & 3                          & 0.09                           & 133.09 \\
		\texttt{D}   & \texttt{a}   &  1  & 1     & 100                   & 40                         & 1                          & 0.1                            & 141.1  \\
		\texttt{D}   & \texttt{b}   &  1  & 1     & 100                   & 40                         & 2                          & 0.11                           & 142.11 \\
		\texttt{D}   & \texttt{c}   &  1  & 1     & 100                   & 40                         & 3                          & 0.12                           & 143.12 \\
		\bottomrule
	\end{tabular}
	\label{table:example1}
\end{table*}

\begin{figure}[htb]
	\centering
	\makeatletter
\newcommand{\MyResizeWidth}{\@ifclassloaded{acmart}{\linewidth}{0.6\textwidth}}
\makeatother
\resizebox{\MyResizeWidth}{!}{%
    \begin{tikzpicture}[]

        \def\sVar{s}
        \def\pVarOne{p1}
        \def\pVarTwo{p2}
        \def\nVar{n}

        \foreach \x/\y/\type in {
            1/1/s,
            2/1/s,
            3/1/s,
            4/1/s,
            5/1/s,
            6/1/s,
            7/1/s,
            8/1/s,
            9/1/s,
            10/1/s,
            11/1/s,
            12/1/s,
            1/2/p1,
            2/2/p1,
            3/2/p1,
            4/3/p1,
            5/3/p1,
            6/3/p1,
            7/4/p1,
            8/4/p1,
            9/4/p1,
            10/5/p1,
            11/5/p1,
            12/5/p1,
            1/6/p2,
            4/6/p2,
            7/6/p2,
            10/6/p2,
            2/7/p2,
            5/7/p2,
            8/7/p2,
            11/7/p2,
            3/8/p2,
            6/8/p2,
            9/8/p2,
            12/8/p2,
            1/9/n,
            2/10/n,
            3/11/n,
            4/12/n,
            5/13/n,
            6/14/n,
            7/15/n,
            8/16/n,
            9/17/n,
            10/18/n,
            11/19/n,
            12/20/n,
        }{

            \ifx\type\sVar
                \node[circle, fill=green!50!black, minimum size=10pt] at (\y, -\x) {};
            \fi
            \ifx\type\pVarOne
                \node[rectangle, fill=orange, minimum size=10pt] at (\y, -\x) {};
            \fi
            \ifx\type\pVarTwo
                \node[rectangle, fill=blue, minimum size=10pt] at (\y, -\x) {};
            \fi
            \ifx\type\nVar
                \node[regular polygon, regular polygon sides=3, fill=red] at (\y, -\x) {};
            \fi

        }
    \end{tikzpicture}
}
	\caption{%
		Sparsity pattern of the Jacobian required for identifying the parameters of \cref{eq:example1} using the example of \cref{table:example1}.
		The circular points correspond to shared parameters, squares correspond to partially-shared parameters, and triangles pertain to non-shared parameters.
	}
	\label{figure:sparsity}
\end{figure}

\subsection{Incorporation into Symbolic Regression}
\label{subsection:incorporation-into-symbolic-regression}

Regardless of the \ac{SR} algorithm this approach is implemented in, the terminal set has to be expanded to accommodate the three types of parameters.
Partially-shared parameters are subdivided based on the categorical variables they reference, as their behavior and the number of individual parameters they contain depend on the specific category.
This may affect several parts of the respective algorithm or implementation.
Here, we briefly describe a selection of the changes required for a \ac{GP} method.

First, the new parameter types can be included during generation of new expressions.
However, they may also be omitted, leaving their introduction to subsequent mutations.
We set the probability of including any but fully-shared parameters into new expression to zero.

Next, a new string representation is required for the different parameter types.
For partially-shared parameters, we use \texttt{C} followed by the category index they refer to (\eg, \texttt{C1} for parameters depending on categorical variable~$1$).
For non-shared parameters depending on the combination of all categories, we use \texttt{CI} (\texttt{I} for interaction).

Prior research \cite{KommendaKronbergerWinklerAffenzellerWagner:2013:1,KronbergerKommendaPrombergerNickel:2018:1} has shown that fewer parameter identification iterations are required if parameter values are cached between generations.
Thus, in our approach, the individual values within a new parameter node are preserved separately, ensuring that genetic operators treat each scalar as a distinct unit.

Mutations which may concern terminals should be adapted for the new terminal set.
We extended standard \enquote{point mutations} to allow nodes to transition between existing and new parameter types.
Also, we modified parameter perturbations to independently adjust each scalar within non-shared or partially-shared parameters.

Finally, because lower levels of sharing increase the number of individual parameters, this should be either directly reflected in the complexity measure or penalized using an additional objective.
We define the number of individual parameters as an additional objective in a multiobjective \ac{GP} approach (see \cref{section:application}).

Many \ac{SR} implementations allow reusing parts of the expression within itself, often referred to as directed acyclic graphs.
This is particularly useful for shared expressions, as it allows a single shared parameter to be referenced in multiple locations, potentially reducing the total count of individual parameters.

\subsection{More Categorical Variables and Limitations}
\label{section:more-categorical-variables}

The concept is explained and demonstrated for two categories, but is generalizable to arbitrarily many categories with increasingly more levels of intermediate sharing.
For two categorical variables, the terminal set is extended by four new types of parameters.
This increases the search space and does thus impact the search efficiency.
However, the number of additional parameter types increases exponentially with the number of categories.
Thus, including all emerging, intermediate sharing types does not scale well with the number of categorical variables.

\section{Related Work}
\label{section:related-work}

The method developed by Kronberger~\etal~\cite{KronbergerKommendaPrombergerNickel:2018:1} is closest to what we propose in this paper.
Focusing on finding common behavior across similar systems, they introduce one categorical variable using category-value-specific parameters in the manner already discussed.
Both fully-shared and category-value-specific parameters are allowed.
Multiple categorical variables are mentioned, but are deemed unfeasible, as interactions could not be accounted for and datasets would become too sparse.
They also compare this varying-parameter-sets-approach to one-hot-encoding, which is another method for including categorical variables.
One-hot-encoding is deemed less interpretable because the categorical influences are distributed across the resulting expressions.

In \cite{LohSchneegassTian:2019:1,TenachiIbataFrancoisDiakogiannis:2024:1,RusseilOlivettiDeFrancaMalanchevBurlacuIshidaLerouxMichelinMoinardGangler:2024:1,KronbergerKommendaPrombergerNickel:2018:1} and the implementations \cite{OlivettiDeFrancaKronberger:2025:2,Cranmer:2020:1}, various approaches are introduced each of which incorporates only one categorical variable.
While \cite{RusseilOlivettiDeFrancaMalanchevBurlacuIshidaLerouxMichelinMoinardGangler:2024:1,OlivettiDeFrancaKronberger:2025:2,LohSchneegassTian:2019:1} allow for category-value-specific parameters only, \cite{KronbergerKommendaPrombergerNickel:2018:1,TenachiIbataFrancoisDiakogiannis:2024:1,Cranmer:2020:1} also include parameters shared across all category values.

Russeil~\etal~\cite{RusseilOlivettiDeFrancaMalanchevMoinardCherrey:2025:1} compares \cite{RusseilOlivettiDeFrancaMalanchevBurlacuIshidaLerouxMichelinMoinardGangler:2024:1,TenachiIbataFrancoisDiakogiannis:2024:1,Cranmer:2020:1,OlivettiDeFrancaKronberger:2025:2} using a common benchmark with real-world problems containing one categorical variable.
They discuss benefits and drawbacks, and offer recommendations for future research in this promising field.
The option of using more than one categorical variable is briefly mentioned in \cite{RusseilOlivettiDeFrancaMalanchevMoinardCherrey:2025:1}, but they do not discuss adapting their methods to accommodate for them.

\section{Application}
\label{section:application}

To demonstrate the proposed approach, first, we further expand on the previously introduced example \cref{eq:example1} with the two categories~$U$ and~$L$.
The presented approach may be uniquely useful for cases, where there are very few data for some of the category combinations, while there are an abundance of data for the others.
We design an example to gauge the parameter sharing capabilities, and thus, the data sharing and transfer-learning limits in terms of data reduction.
In essence, we seek to understand to what degree data distributed across the category combinations may be used between the categories to correctly learn the parameters for category combinations with very limited data.

\Cref{eq:example1}, along with its parameters, remains the same.
For $v_1$ of each of the twelve category-value combinations, eight data points are randomly and uniformly sampled in the range of $\interval[][]{-20}{20}$, resulting in a total of $96$~data points.
In each step of a \enquote{procession}, a data point is randomly transferred from the training set to the test set.
All individual parameters $p$ are then randomly perturbed with
\begin{equation}
	\label{eq:perturb}
	p_{\textup{perturbed}}
	=
	p + 0.1 \cdot p \cdot r
	\quad
	\text{with }
	r \sim \cN(0,1)
\end{equation}
before they are re-identified and the prediction is compared to the test set.
For one procession, this is repeated until the minimum data requirements are no longer satisfied.
Excerpts of four of a total of $100$~conducted processions are shown in \cref{table:sharing}.

\begin{table*}[htb]
	\centering
	\caption{%
		Excerpts of four processions, each randomly, iteratively moving points from the training dataset to the test dataset, perturbing the parameters, reidentifying them, and determining the mean squared error ($\textup{mse}_{\textup{test}}$) between the predictions and the test dataset.
		The ID is made up of a procession number and the number of total data points, the columns \texttt{Aa} \ldots \texttt{Dc} show the number of data points in each category, and the \enquote{req.} column indicates whether the minimum data requirements are fulfilled.
	}
	\begin{tabular}{lllllllllllllll}
		\toprule
		ID     & \texttt{Aa} & \texttt{Ab} & \texttt{Ac} & \texttt{Ba} & \texttt{Bb} & \texttt{Bc} & \texttt{Ca} & \texttt{Cb} & \texttt{Cc} & \texttt{Da} & \texttt{Db} & \texttt{Dc} & $\textup{mse}_{\textup{test}}$   & req.\ \\
		\midrule
		1:96   & 8           & 8           & 8           & 8           & 8           & 8           & 8           & 8           & 8           & 8           & 8           & 8           & N/A                              & yes   \\
		1:90   & 8           & 7           & 8           & 8           & 8           & 8           & 5           & 8           & 8           & 8           & 6           & 8           & 2e-25                            & yes   \\
		1:60   & 7           & 3           & 5           & 7           & 6           & 4           & 3           & 6           & 6           & 5           & 3           & 5           & 3e-24                            & yes   \\
		1:30   & 2           & 2           & 1           & 3           & 4           & 1           & 1           & 5           & 5           & 4           & 1           & 1           & 2e-22                            & yes   \\
		1:22   & 2           & 1           & 1           & 2           & 3           & 1           & 1           & 2           & 5           & 2           & 1           & 1           & 2e-11                            & yes   \\
		1:21   & 2           & 1           & 1           & 2           & 3           & 1           & 1           & 2           & 5           & 1           & 1           & 1           & 1e6                              & no    \\
		\midrule
		2:20   & 1           & 1           & 2           & 1           & 4           & 1           & 2           & 1           & 1           & 2           & 2           & 2           & 7e-10                            & yes   \\
		2:19   & 1           & 1           & 2           & 1           & 4           & 1           & 2           & 1           & 1           & 1           & 2           & 2           & 1e4                              & no    \\
		\midrule
		3:26   & 3           & 1           & 1           & 1           & 4           & 1           & 1           & 4           & 5           & 1           & 3           & 1           & 3e-23                            & yes   \\
		3:25   & 2           & 1           & 1           & 1           & 4           & 1           & 1           & 4           & 5           & 1           & 3           & 1           & 1e-2                             & no    \\
		\midrule
		4:48   & 5           & 6           & 3           & 5           & 6           & 7           & 4           & 5           & 3           & 1           & 1           & 2           & 2e-23                            & yes   \\
		4:47   & 5           & 6           & 3           & 5           & 6           & 7           & 4           & 5           & 3           & 1           & 1           & 1           & 6e3                              & no    \\
		\bottomrule
	\end{tabular}
	\label{table:sharing}
\end{table*}

As can be seen in \cref{table:sharing}, the test data are predicted well, despite very few data points in many of the category-value combinations.
For many category-value combinations, only one data point is required, as long as there are sufficient data in other category-value combinations.
For example, in ID~3:26, seven out of twelve category-value combinations require only a single data point.

In this example, there are $20$~individual parameters, which act like $48$~parameters.
Thus, in the best case, as can be seen in procession~$2$, only a minimum of $20$~data points are required to identify the parameters.
However, as can be seen in processions~$1$, $3$, and $4$, the minimum data requirements are no longer fulfilled despite the availability of more than $20$~points.
This stems from the fact that data points only contribute to the identification of parameters they specifically reference.

For each non-shared parameter, one point is required in each category-value combination.
For a partially-shared parameter, at least one additional point is required for each value of the category upon which it depends.
And finally, one additional point in any category-value combination is required to identify a shared parameter.

Whether the data distribution across the category-value combinations is sufficient according to those conditions is indicated in the \enquote{req.} column of \cref{table:sharing}.
As evident in the table, as soon as the minimum data requirements are no longer satisfied, the test data can no longer be predicted well, which is reflected in all of the $100$~conducted processions.
Of course, sufficient data for the parameter identification alone does not guarantee a successful refitting, as it also depends on the distribution of the $v_1$ values in the training dataset.
To conclude this example, although there are four parameters in the expression, the parameters can be identified correctly for category combinations containing only one point.
This highlights the strengths and limitations of our approach's transfer learning capabilities.

The proposed approach incorporates multiple categorical variables in a novel manner.
Although we expect a wide applicability of the proposed approach to many scientific and engineering domains, publicly available datasets containing multiple categorical variables remain scarce.
The methods introduced by prior studies are limited to one categorical variable, and most of their examples do only contain one.
One exception is the astrophysics example introduced by \cite{RusseilOlivettiDeFrancaMalanchevBurlacuIshidaLerouxMichelinMoinardGangler:2024:1,RusseilOlivettiDeFrancaMalanchevMoinardCherrey:2025:1}, which we utilize for the present study.

This dataset contains two categorical variables, which the authors flattened to one by considering their combination as a single categorical variable.
It describes the change of radiation flux of supernovae (three datasets) over time using two types of photometric filters (two bands), which results in three times two (equals six) unique category-value combinations.

\begin{table*}[h]
	\centering
	\caption{Expressions reported in \cite{RusseilOlivettiDeFrancaMalanchevBurlacuIshidaLerouxMichelinMoinardGangler:2024:1} for the supernovae dataset along with their $R^2$-values, and the number of individual parameter. In their study, only non-shared parameters ($C_{\textup{i}}$) are used; in this context, each parameter node represents six individual scalar values (one per category combination).}
	\begin{tabular}{lrr}
		\toprule
		Expression                                                                                                        & $R^2$ & \# of individual parameters \\
		\midrule
		$e^{-C_{\textup{i},1}t} \cdot (C_{\textup{i},2} - e^{-C_{\textup{i},3}t})$                                        & 0.990 & 18 \\
		\midrule
		$\frac{C_{\textup{i},1}}{(C_{\textup{i},2} \cdot e^{C_{\textup{i},3}t} + e^{-C_{\textup{i},4}t})}$                & 0.987 & 24 \\
		\midrule
		$\frac{C_{\textup{i},1}^{C_{\textup{i},2}t}}{C_{\textup{i},3}t + (-C_{\textup{i},4}t + e^{C_{\textup{i},5}t})^2}$ & 0.992 & 30 \\
		\bottomrule
	\end{tabular}
	\label{table:supernovae:multiview}
\end{table*}

In \cite{RusseilOlivettiDeFrancaMalanchevBurlacuIshidaLerouxMichelinMoinardGangler:2024:1}, the authors show three resulting expressions for this problem with three, four, and five parameters and $R^2$-values of~$0.99$, $0.987$, and $0.992$, respectively.
However, as they only contain the present-studies-equivalent of non-shared parameters, each of those contribute six individual parameters, resulting in $18$, $24$, and $30$~individual parameters.
The expressions and the relevant information are shown in \cref{table:supernovae:multiview}.

We implement the proposed approach on the basis of the open-source \ac{SR} package \tisr \cite{MartinekFrotscherRichterHerzog:2023:1,Martinek:2023:1}.
\tisr leverages NSGA-II to provide a multi-objective approach to \ac{GP}, featuring a highly modular architecture for adding custom objectives and configurations.
To evaluate different approaches to handling categorical variables, we implement five variants inside the same algorithmic framework.
Four of the five variants (\texttt{conventional}, \texttt{separate}, \texttt{flattened}, \texttt{predictive}) represent established methodologies, while one (\texttt{novel}) is novel.

The first variant, \texttt{conventional}, utilizes standard \ac{SR} on the entire dataset, effectively ignoring categorical distinctions.
Within our framework, this corresponds to generating expressions using only fully-shared parameters.
At the opposite extreme, the \texttt{separate} variant treats each category-value combination independently.
The two categories are collapsed into one by considering the combinations of category values as a single categorical variable.
While the expressions share a functional form, their parameters are entirely separate and thus non-shared.
This variant aligns with methods proposed by \cite{RusseilOlivettiDeFrancaMalanchevBurlacuIshidaLerouxMichelinMoinardGangler:2024:1,OlivettiDeFrancaKronberger:2025:2,LohSchneegassTian:2019:1}.
The \texttt{flattened} variant also uses only a single flattened categorical variable, but allows both fully-shared and non-shared parameters, consistent with the approaches in \cite{TenachiIbataFrancoisDiakogiannis:2024:1,Cranmer:2020:1}.
The \texttt{predictive} variant exclusively utilizes full and partial parameter sharing, omitting non-shared parameters.
This variant is especially useful for zero-shot generalization; by excluding parameters tied to specific category interactions, the resulting expressions can be applied to out-of-distribution category combinations entirely absent from the training dataset.
The approach \cite{KronbergerKommendaPrombergerNickel:2018:1} is aligned with this approach.
However, for one categorical variable, their approach also allows realizing the \texttt{flattened} variant.
Lastly, the \texttt{novel} variant incorporates all three parameter-sharing types discussed in the present work.
The variants and their sharing modalities are summarized in \cref{table:sharing:modalities}.

\begin{table}[htb]
    \centering
    \caption{Summary of the algorithmic variants and their sharing modalities.}
    \begin{tabular}{llr}
        \toprule
        label                 & sharing modalities                            & \# of categ. var. \\
        \midrule
        \texttt{conventional} & fully-shared                                  & 0 \\
        \texttt{separate}     & non-shared                                    & 1 \\
        \texttt{flattened}    & fully-shared \& non-shared                    & 1 \\
        \texttt{predictive}   & fully-shared \& partially-shared              & 2 \\
        \texttt{novel}        & fully-shared, partially-shared, \& non-shared & 2 \\
        \bottomrule
    \end{tabular}
    \label{table:sharing:modalities}
\end{table}

\begin{figure}[htp]
	\centering
	\begin{subfigure}[b]{0.8\textwidth}
		\centering
		\includegraphics[width=1.0\textwidth]{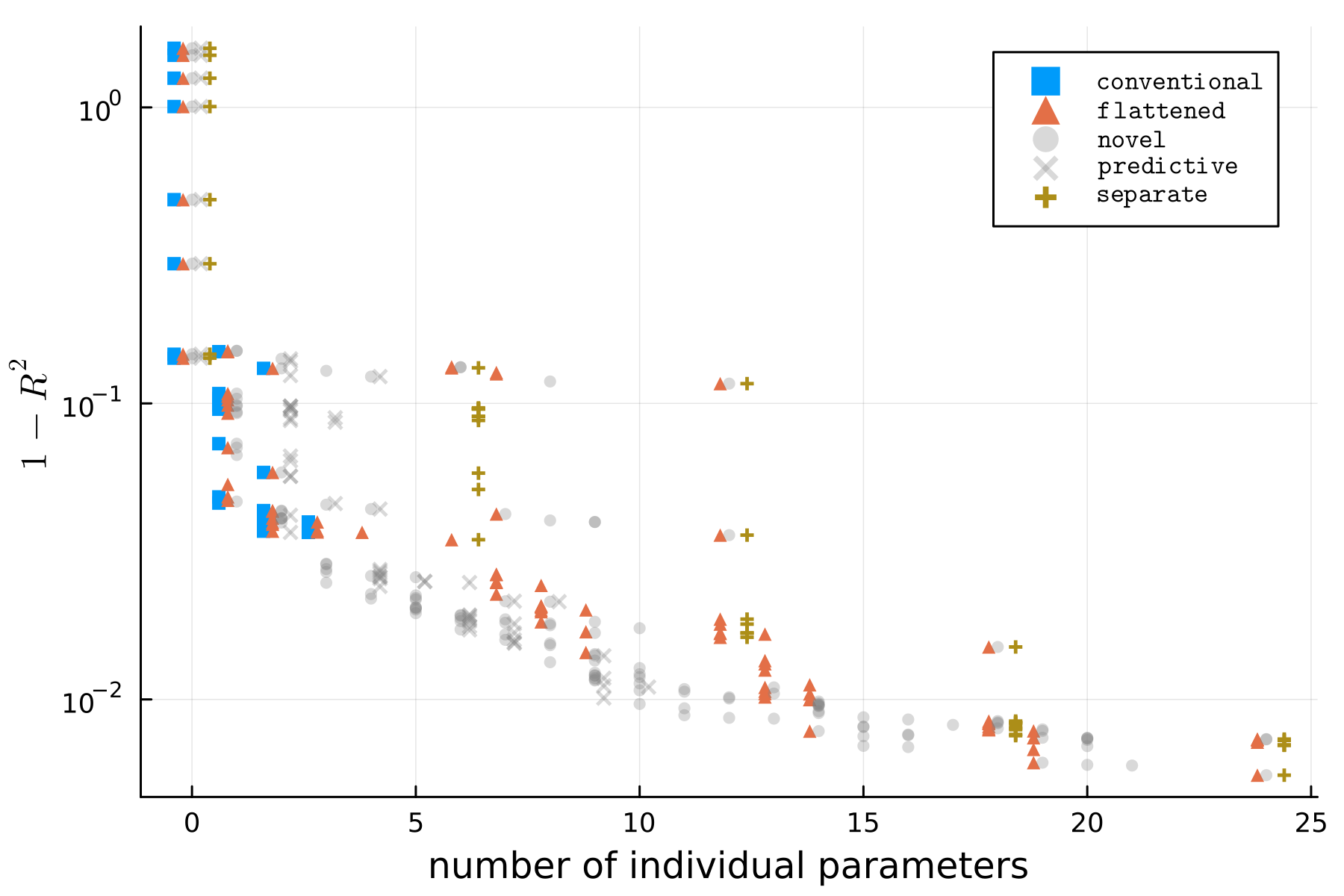}
	\end{subfigure}
	\begin{subfigure}[b]{0.8\textwidth}
		\centering
		\includegraphics[width=1.0\textwidth]{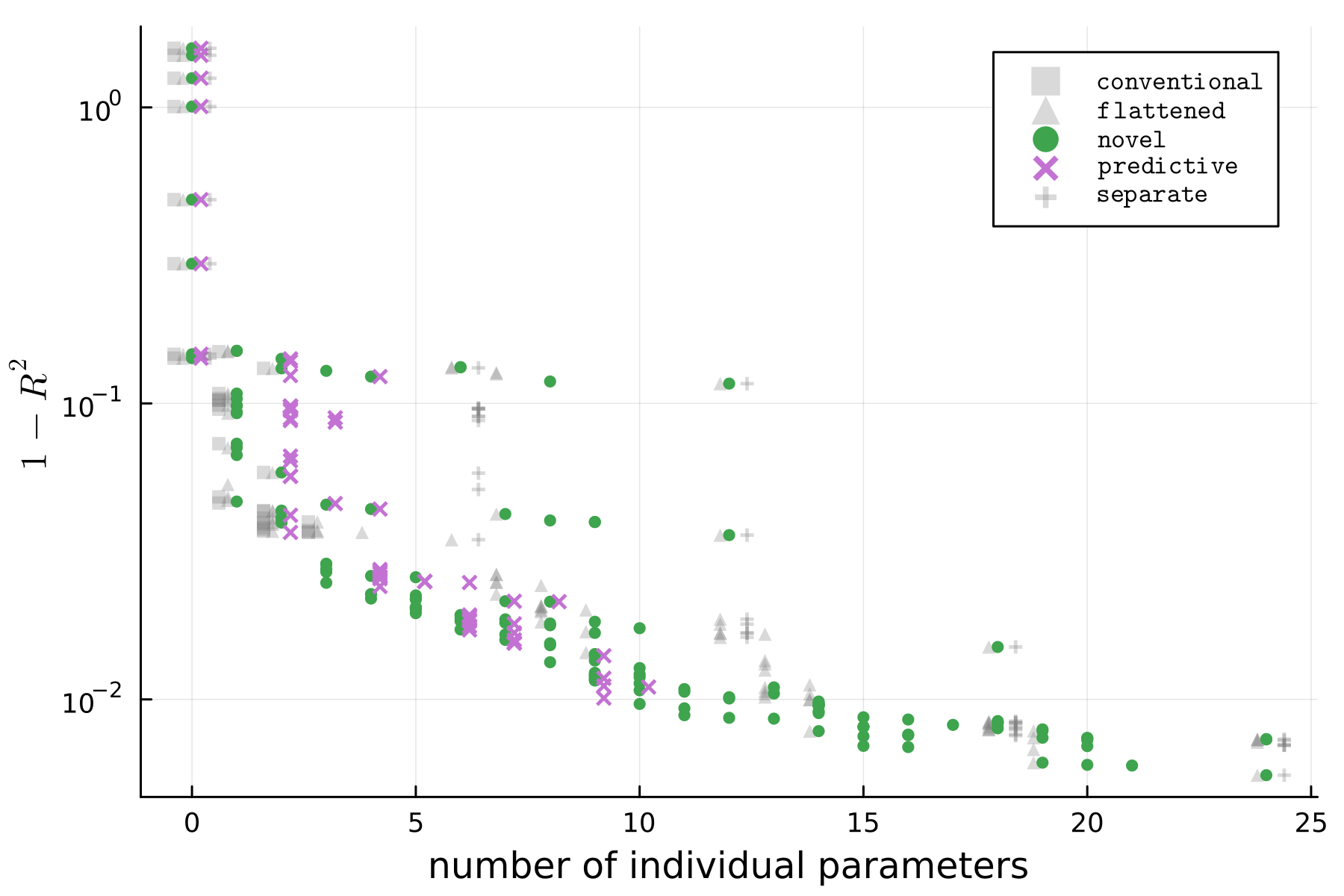}
	\end{subfigure}
	\caption{Performance comparison of five categorical variable handling variants, plotted by $1 - R^2$ versus the number of individual parameters. Data points represent expressions from five $15$-minute runs per variant. To improve clarity, a \enquote{niching} process is applied to each variant independently: $1 - R^2$ values are rounded to three significant digits, and only unique solutions are retained. A horizontal jitter is applied to the x-axis to prevent marker overlap. Both subplots display the full dataset, with specific variants grayed out in each to facilitate direct comparison.}
	\label{figure:pareto}
\end{figure}

For all experiments, we adopt the dataset and preprocessing of \cite{RusseilOlivettiDeFrancaMalanchevBurlacuIshidaLerouxMichelinMoinardGangler:2024:1}, and choose the function set as
\begin{equation*}
	\set{%
		\texttt{+}
		,\,
		\texttt{-}
		,\,
		\texttt{*}
		,\,
		\texttt{/}
		,\,
		\texttt{\^{}}
		,\,
		\texttt{exp}
		,\,
		\texttt{log}
		,
		\texttt{square}
		,
		\texttt{sqrt}
	}
	.
\end{equation*}
The selection objectives are $1 - R^2$, the complexity (number of operators and operands), and the number of individual parameters.
The maximum allowed complexity is limited to~$12$ to allow direct comparisons to \cite{RusseilOlivettiDeFrancaMalanchevBurlacuIshidaLerouxMichelinMoinardGangler:2024:1}.
Each of the five variants is executed five times, with each run lasting $15$~minutes.
The $1 - R^2$ and number of individual parameters of the expressions found using each variant are shown in \cref{figure:pareto}.

As can be seen in \cref{figure:pareto}, the \texttt{conventional} variant only produces solutions with a small number of individual parameters.
Among the tested variants, its expressions are Pareto-optimal only for two or fewer parameters.
The \texttt{separate} variant generates expressions with parameter counts in multiples of six; this is because it relies exclusively on non-shared parameters, each of which contributes six individual parameters to the total.
Compared to other variants, \texttt{separate} only reaches the Pareto frontier at zero, $18$, and $24$~individual parameters.
At the zero-parameter baseline, all studied approaches perform identically.
For $18$ and $24$~parameters, approaching upper extreme of number of parameters of this study, the performance of \texttt{separate} and \texttt{flattened} converges.
However, because \texttt{flattened} can utilize shared parameters, it possesses greater flexibility in allocating its parameter budget.
It identifies superior solutions at six and twelve individual parameters compared to \texttt{separate}, and provides more granular trade-offs at counts that are not multiples of six.

A similar improvement is observed when comparing the \texttt{separate} and \texttt{novel} variants.
The \texttt{novel} approach leverages greater degrees of freedom by optimizing both the allocation of individual parameters and the choice of sharing modality.
Consequently, its solutions lie almost exclusively on the Pareto frontier.
Across a wide range of parameter counts, it produces expressions on par with the best-performing variants, while often uniquely identifying the best trade-offs.
Finally, the \texttt{predictive} variant produces expressions limited to ten or fewer parameters.
However, it successfully reaches the Pareto frontier at three, six, seven, and nine individual parameters.

\begin{table*}[htb]
	\centering
	\caption{%
		Excerpt of expressions found for the supernovae dataset \cite{RusseilOlivettiDeFrancaMalanchevBurlacuIshidaLerouxMichelinMoinardGangler:2024:1} along with their $R^2$-values, and the number of individual parameters.
		The non-shared parameters are denoted by $C_{\textup{i}}$ (each carrying six individual parameters), the partially-shared parameters depending on the type of band are shown as $C_{\textup{band}}$ (each carrying two individual parameters), and the partially-shared parameters depending on the dataset are shown as $C_{\textup{dataset}}$ (each carrying three individual parameters).
	}
	\begin{tabular}{lrr}
		\toprule
		Expression                                                                                                                             & $R^2$ & \# of individual parameters \\
		\midrule
		$C_{\textup{i},1}^{\left( C_{\textup{i},2} - t \right)^{2} \cdot \left( 0.134 + C_{\textup{band}}^{t} \right)}$                        & 0.993 & 15 \\
		\midrule
		$0.998^{\left( C_{\textup{i},1} + 0.961^{t} \right)^{2} \cdot \left( C_{\textup{i},2} - t \right)^{2}}$                                & 0.991 & 14 \\
		\midrule
		$\left( C_{\textup{i},1} + C_{\textup{i},2} \cdot C_{\textup{band}}^{t} \right)^{t}$                                                   & 0.99  & 14 \\
		\midrule
		$\left(  - C_{\textup{band},2} + C_{\textup{dataset}} \cdot C_{\textup{band},1}^{t} \right)^{C_{\textup{band},3} \cdot t}$             & 0.989 & 9 \\
		\midrule
		$C_{\textup{band},1}^{\left( C_{\textup{band},2} + \frac{C_{\textup{dataset}}}{C_{\textup{band},3} - t} \right)^{2}}$                  & 0.986 & 9 \\
		\midrule
		$C_{\textup{band},1}^{\left( 0.116 + \frac{ - C_{\textup{dataset}} + t}{C_{\textup{band},2} - t} \right)^{2}}$                         & 0.985 & 8 \\
		\bottomrule
	\end{tabular}
	\label{table:supernovae}
\end{table*}

Lastly, an excerpt of the expressions found using the variants \texttt{novel} and \texttt{predictive} are shown in \cref{table:supernovae}.
Although the \texttt{novel} variant has the flexibility to use any sharing type, it also identifies Pareto-optimal expressions that omit non-shared parameters, particularly when the search is incentivized by the parameter-count objective.
These solutions are highly desirable as they provide the best trade-off between accuracy and the ability to generalize to unseen category combinations.

All the shown expressions have fewer individual parameters than the ones reported by \cite{RusseilOlivettiDeFrancaMalanchevBurlacuIshidaLerouxMichelinMoinardGangler:2024:1}, while they retain a similar $R^2$-value.
For example, the expression in the third row of \cref{table:supernovae} uses one parameter depending on the type of band, and two non-shared parameters, which results in only $14$~individual parameters.
Although it uses four fewer individual parameters, its $R^2$-value is the same as that of the three-parameter expression from \cite{RusseilOlivettiDeFrancaMalanchevBurlacuIshidaLerouxMichelinMoinardGangler:2024:1} ($R^2 = 0.99$), shown in the first row of \cref{table:supernovae:multiview}.
Comparing the best expressions of the present study and \cite{RusseilOlivettiDeFrancaMalanchevBurlacuIshidaLerouxMichelinMoinardGangler:2024:1}, the proposed approach requires half of the parameters ($15$ versus $30$) to achieve a similar $R^2$ value ($R^2 = 0.993$ versus $0.992$).

An additional benefit of the proposed approach is that more information is can be extracted about the problem and the categories.
By inspecting the resulting expressions, domain experts can distinguish between effects that are category-agnostic, category-specific, or unique to particular category-value interactions.

\section{Conclusion}
\label{section:conclusion}

This paper presents a novel framework for incorporating multiple categorical variables into \ac{SR}.
By introducing fully-shared, partially-shared, and non-shared parameters, our approach significantly extends the flexibility of \ac{SR} in the presence of multiple categorical variables.
These sharing modalities allow the model to simultaneously capture category-agnostic trends, isolate category-specific effects, and account for unique category-value interactions.
This enables the identification of additional problem structure, reduces the number of individual parameters, and lowers data requirements.
Through a synthetic benchmark, we characterize the limits of data sharing and transfer learning within this framework.
The results show that the proposed sharing mechanisms dramatically reduce per-combination data requirements, provided that sufficient global data exist to identify the shared parameters.
This framework can be implemented in any \ac{SR} architecture that allows for the expansion of the terminal set with new parameter types and the necessary adaptations to identify those parameters.

We implement the framework in a multiobjective \ac{GP}-based \ac{SR} algorithm and demonstrate it on a real-world astrophysics dataset used in a related study \cite{RusseilOlivettiDeFrancaMalanchevBurlacuIshidaLerouxMichelinMoinardGangler:2024:1,RusseilOlivettiDeFrancaMalanchevMoinardCherrey:2025:1}.
Within the same framework, we implement five algorithmic variants that incorporate categorical variables in different ways.
Four variants align with prior methods, while the remaining one is based on our proposed method.
Comparative analysis shows that our proposed variant consistently outperforms established methods by identifying a more diverse Pareto frontier of solutions with superior trade-offs between fit quality and parameter count.
The proposed approach will be merged into the publicly available branch of \tisr in the future.

% Insert the appendix.
\appendix

% Insert the bibliography.
\printbibliography

\end{document}